\documentclass[a4paper,twoside]{article}

\usepackage{epsfig}
\usepackage{subcaption}
\usepackage{calc}
\usepackage{amssymb}
\usepackage{amstext}
\usepackage{amsmath}
\usepackage{amsthm}
\usepackage{multicol}
\usepackage{pslatex}
\usepackage{apalike}
\usepackage{multirow}
\usepackage{transparent}
\usepackage{url}
\usepackage{float}
\usepackage{CJKutf8}
\usepackage{graphicx}
\usepackage{xcolor,colortbl}
\definecolor{Gray}{gray}{0.85}
\definecolor{cya}{rgb}{0.88,1,1}
\definecolor{gre}{rgb}{0.55, 0.71, 0.0}
\definecolor{apr}{rgb}{0.98, 0.81, 0.69}
\definecolor{yel}{rgb}{0.89, 0.82, 0.04}
\definecolor{lav}{rgb}{0.75, 0.58, 0.89}
\definecolor{red}{rgb}{0.8, 0.36, 0.36}
\definecolor{whi}{rgb}{1,1,1}

\usepackage{SCITEPRESS}     % Please add other packages that you may need BEFORE the SCITEPRESS.sty package.

\begin{document}

\title{Data Mining in Clinical Trial Text: Transformers for Classification and Question Answering Tasks}

\author{\authorname{Lena~Schmidt\sup{1}\sup{2}\orcidAuthor{0000-0003-0709-8226}, Julie~Weeds\sup{2}\orcidAuthor{0000-0002-3831-4019} and Julian~P.~T.~Higgins\sup{1}\orcidAuthor{0000-0002-8323-2514}}
\affiliation{\sup{1}University of Bristol, Bristol Medical School, 39 Whatley Road, BS82PS Bristol, UK}
\affiliation{\sup{2}University of Sussex, Department of Informatics, BN19QJ Brighton, UK}
\email{\{lena.schmidt, julian.higgins\}@bristol.ac.uk, j.e.weeds@sussex.ac.uk}
}


\abstract{This research on data extraction methods applies recent advances in natural language processing to evidence synthesis based on medical texts. Texts of interest include abstracts of clinical trials in English and in multilingual contexts. The main focus is on information characterized via the Population, Intervention, Comparator, and Outcome (PICO) framework, but data extraction is not limited to these fields. Recent neural network architectures based on transformers show capacities for transfer learning and increased performance on downstream natural language processing tasks such as universal reading comprehension, brought forward by this architecture's use of contextualized word embeddings and self-attention mechanisms. This paper contributes to solving problems related to ambiguity in PICO sentence prediction tasks, as well as highlighting how annotations for training named entity recognition systems are used to train a high-performing, but nevertheless flexible architecture for question answering in systematic review automation. Additionally, it demonstrates how the problem of insufficient amounts of training annotations for PICO entity extraction is tackled by augmentation. All models in this paper were created with the aim to support systematic review (semi)automation. They achieve high F1 scores, and demonstrate the feasibility of applying transformer-based classification methods to support data mining in the biomedical literature.}

\onecolumn \maketitle \normalsize \setcounter{footnote}{0} \vfill

\section{\uppercase{Introduction}}
\label{sec:introduction}

\noindent Systematic reviews (SR) of randomized controlled trials (RCTs) are regarded as the gold standard for providing information about the effects of interventions to healthcare practitioners, policy makers and members of the public. The quality of these reviews is ensured through a strict methodology that seeks to include all relevant information on the review topic \cite{coc1}.~\\

A SR, as produced by the quality standards of Cochrane, is conducted to appraise and synthesize all research for a specific research question, therefore providing access to the best available medical evidence where needed \cite{coc5}. The research question is specified using the PICO (population; intervention; comparator; outcomes) framework. The researchers conduct very broad literature searches in order to retrieve every piece of clinical evidence that meets their review's inclusion criteria, commonly all RCTs of a particular healthcare intervention in a specific population. In a search, no piece of relevant information should be missed. In other words, the aim is to achieve a recall score of one. This implies that the searches are broad \cite{coc6}, and authors are often left to screen a large number of abstracts manually in order to identify a small fraction of relevant publications for inclusion in the SR \cite{Bor17}.

The number of RCTs is increasing, and with it increases the potential number of reviews and the amount of workload that is implied for each. Research on the basis of PubMed entries shows that both the number of publications and the number of SRs increased rapidly in the last ten years \cite{fon18}, which is why acceleration of the systematic reviewing process is of interest in order to decrease working hours of highly trained researchers and to make the process more efficient.\\~

In this work, we focus on the detection and annotation of information about the PICO elements of RCTs described in English PubMed abstracts. In practice, the comparators involved in the C of PICO are just additional interventions, so we often refer to PIO (populations; interventions; outcomes) rather than PICO. Focus points for the investigation are the problems of ambiguity in labelled PIO data, integration of training data from different tasks and sources and assessing our model's capacity for transfer learning and domain adaptation.

Recent advances in natural language processing (NLP) offer the potential to be able to automate or semi-automate the process of identifying information to be included in a SR.  For example, an automated system might attempt to PICO-annotate large corpora of abstracts, such as RCTs indexed on PubMed, or assess the results retrieved in a literature search and predict which abstract or full text article fits the inclusion criteria of a review. Such systems need to be able to classify and extract data of interest. We show that transformer models perform well on complex data-extraction tasks. Language models are moving away from the semantic, but static representation of words as in Word2Vec \cite{mik13}, hence providing a richer and more flexible contextualized representation of input features within sentences or long sequences of text. 

The rest of this paper is organized as follows. The remainder of this section introduces related work and the contributions of our work. Section 2 describes the process of preparing training data, and introduces approaches to fine-tuning for sentence classification and question answering tasks. Results are presented in section 3, and section 4 includes a critical evaluation and implications for practice.

\subsection{Tools for SR automation and PICO classification}

The website \url{systematicreviewtools.com} \cite{sys19} lists 36 software tools for study selection to date. Some tools are intended for organisational purposes and do not employ PICO classification, such as Covidence \cite{cov19}. The tool Rayyan uses support vector machines \cite{ray16}. RobotReviewer uses neural networks, word embeddings and recently also a transformer for named entity recognition (NER) \cite{Mar17}. Question answering systems for PICO data extraction exist based on matching words from knowledge bases, hand-crafted rules and na\"ive Bayes classification, both on entity and sentence level \cite{dem05}, \cite{niu03}, but commonly focus on providing information to practicing clinicians rather than systematic reviewers \cite{von15}. 

In the following we introduce models related to our sentence and entity classification tasks and the data on which our experiments are based. We made use of previously published training and testing data in order to ensure comparability between models.\\

\subsection{Sentence classification data}

%%%%5DATASET 1 & comparator model
In the context of systematic review (semi)automation, sentence classification can be used in the screening process, by highlighting relevant pieces of text. A long short-term memory (LSTM) neural network trained with sentences of structured abstracts from PubMed was published in 2018 \cite{jin18}. It uses a pre-trained Word2Vec embedding in order to represent each input word as a fixed vector. Due to the costs associated with labelling, its authors acquired sentence labels via automated annotation. Seven classes were assigned on the basis of structured headings within the text of each abstract. Table~\ref{tab:classes} provides an overview of class abbreviations and their meaning.\footnote{Intervention and comparator are commonly combined in PICO annotation tasks, please see original publication for more details \cite{jin18} }In the following we refer to it as the PubMed data.  

The LSTM itself yields impressive results with F1 scores for annotation of up to 0.85 for PIO elements, it generalizes across domains and assigns one label per sentence. We were able to confirm these scores by replicating a local version of this model.

\begin{table}[ht]
\caption{Classes for the sentence classification task.}\label{tab:classes} \centering
\begin{tabular}{l|l}
  \hline
  Class & Abbreviation \\
  \hline\hline
  P&Population\\\hline
     I&Intervention, Comparator\\\hline
     O&Outcome\\\hline
     A&Aim\\\hline
     M&Method\\\hline
     R&Result\\\hline
     C&Conclusion\\\hline
  
\end{tabular}
\end{table}

\subsection{Question answering data}
%%%%5DATASET 2
\subsubsection{SQuAD}
The Stanford Question Answering Dataset (SQuAD) is a reading-comprehension dataset for machine learning tasks. It contains question contexts, questions and answers and is available in two versions. The older version contains only questions that can be answered based on the given context. In its newer version, the dataset also contains questions which can not be answered on the basis of the given context. The SQuAD creators provide an evaluation script, as well as a public leader board to compare model performances \cite{raj16}.  
\subsubsection{Ebm-nlp}
%%%%5DATASET 2
In the PICO domain, the potential of NER was shown by Nye and colleagues in using transformers, as well as LSTM and conditional random fields. In the following, we refer to these data as the ebm-nlp corpus. \cite{ny18}. The ebm-nlp corpus provided us with 5000 tokenized and annotated RCT abstracts for training, and 190 expert-annotated abstracts for testing. Annotation in this corpus include PIO classes, as well as more detailed information such as age, gender or medical condition. We adapted the human-annotated ebm-nlp corpus of abstracts for training our QA-BERT question answering system.

\subsection{Introduction to transformers}

In the following, the bidirectional encoder representations from transformers (BERT) architecture is introduced \cite{dev18}. This architecture's key strengths are rooted in both feature representation and training. A good feature representation is essential to ensure any model's performance, but often data sparsity in the unsupervised training of embedding mechanisms leads to losses in overall performance. By employing a word piece vocabulary, BERT eliminated the problem of previously unseen words. Any word that is not present in the initial vocabulary is split into a sub-word vocabulary. Especially in the biomedical domain this enables richer semantic representations of words describing rare chemical compounds or conditions. A relevant example is the phrase ’two drops of ketorolac tromethamine’, where the initial three words stay intact, while the last words are tokenized to ’ket’, ’\#oro’, ’\#lac’, ’tro’, ’\#meth’, ’\#amine’, hence enabling the following model to focus on relevant parts of the input sequence, such as syllables that indicate chemical compounds. When obtaining a numerical representation for its inputs, transformers apply a ’self-attention’ mechanism, which leads to a contextualized representation of each word with respect to its surrounding words.     

BERT's weights are pre-trained in an unsupervised manner, based on large corpora of unlabelled text and two pre-training objectives. To achieve bidirectionality, its first pre-training objective includes prediction of randomly masked words. Secondly, a next-sentence prediction task trains the model to capture long-term dependencies. Pre-training is computationally expensive but needs to be carried out only once before sharing the weights together with the vocabulary. Fine-tuning to various downstream tasks can be carried out on the basis of comparably small amounts of labelled data, by changing the upper layers of the neural network to classification layers for different tasks. 

SCIBERT is a model based on the BERT-base architecture, with further pre-trained weights based on texts from the Semantic Scholar search engine \cite{sem19}. We used these weights as one of our three starting points for fine-tuning a sentence classification architecture \cite{Bel19}. Furthermore, BERT-base (uncased) and Bert multilingual (cased, base architecture) were included in the comparison \cite{dev18}.

%
%
%%%%made new subsection in order to explain the background to our contributions
%%
\subsection{Weaknesses in the previous sentence classification approach}
In the following, we discuss weaknesses in the PubMed data, and LSTM models trained on this type of labelled data. LSTM architectures commonly employ a trimmed version of Word2Vec embeddings as embedding layer. In our case, this leads to 20\% of the input data being represented by generic `Unknown' tokens. These words are missing because they occur so rarely  that no embedding vector was trained for them. Trimming means that the available embedding vocabulary is then further reduced to the known words of the training, development and testing data, in order to save memory and increase speed. The percentage of unknown tokens is likely to increase when predicting on previously unseen and unlabelled data. We tested our locally trained LSTM on 5000 abstracts from a study-based register \cite{Sho19} and found that 36\% of all unique input features did not have a known representation.     

In the case of the labelled training and testing data itself, automatic annotation carries the risk of producing wrongly labelled data. But it also enables the training of neural networks in the first place because manual gold standard annotations for a project on the scale of a LSTM are expensive and time-consuming to produce. As we show later, the automated annotation technique causes noise in the evaluation because as the network learns, it can assign correct tags to wrongly labelled data. We also show that sentence labels are often ambiguous, and that the assignment of a single label limits the quality of the predictions for their use in real-world reviewing tasks.

We acknowledge that the assignment of classes such as `Results' or `Conclusions' to sentences is potentially valuable for many use-cases. However, those sentences can contain additional information related to the PICO classes of interest. In the original LSTM-based model the A, M, R, and C data classes in Table~\ref{tab:classes} are utilized for sequence optimization, which leads to increased classification scores. Their potential PICO content is neglected, although it represents crucial information in real-world reviewing tasks. 

A general weakness of predicting labels for whole sentences is the practical usability of the predictions. We will show sentence highlighting as a potential use-case for focusing reader's attention to passages of interest. However, the data obtained through this method are not fine-grained enough for usage in data extraction, or for the use in pipelines for automated evidence synthesis. Therefore, we expand our experiments to include QA-BERT, a question-answering model that predicts the locations of PICO entities within sentences.   

\subsection{Contributions of this research}
In this work we investigate state-of-the-art methods for language modelling and sentence classification. Our contributions are centred around developing transformer-based fine-tuning approaches tailored to SR tasks. We compare our sentence classification with the LSTM baseline and evaluate the biggest set of PICO sentence data available at this point \cite{jin18}. We demonstrate that models based on the BERT architecture solve problems related to ambiguous sentence labels by learning to predict multiple labels reliably. Further, we show that the improved feature representation and contextualization of embeddings lead to improved performance in biomedical data extraction tasks. These fine-tuned models show promising results while providing a level of flexibility to suit reviewing tasks, such as the screening of studies for inclusion in reviews. By predicting on multilingual and full text contexts we showed that the model's capabilities for transfer learning can be useful when dealing with diverse, real-world data.

In the second fine-tuning approach, we apply a question answering architecture to the task of data extraction. Previous models for PICO question answering relied on vast knowledge bases and hand-crafted rules. Our fine-tuning approach shows that an abstract as context, together with a combination of annotated PICO entities and SQuAD data can result in a system that outperforms contemporary entity recognition systems, while retaining general reading comprehension capabilities.

\section{\uppercase{Methodology}}
\subsection{Feature representation and advantages of contextualization}

A language processing model's performance is limited by its capability of representing linguistic concepts numerically. In this preliminary experiment, we used the PubMed corpus for sentence classification to show the quality of PICO sentence embeddings retrieved from BERT. We mapped a random selection of 3000 population, intervention, and outcome sentences from the PubMed corpus to BERT-base uncased and SCIBERT. This resulted in each sentence being represented by a fixed length vector of 768 dimensions in each layer respectively, as defined by the model architecture's hidden size. These vectors can be obtained for each of the network's layers, and multiple layers can be represented together by concatenation and pooling.\footnote{A more detailed explanation is given by the BERT authors in their paper on pre-training \cite{dev18b}, and in the Bert-as-service GitHub repository \cite{xia18}.} We used the t-distributed Stochastic Neighbour Embedding (t-SNE) algorithm to reduce each layer-embedding into two-dimensional space, and plotted the resulting values. Additionally, we computed adjusted rand scores in order to evaluate how well each layer (or concatenation thereof, always using reduce\_mean pooling) represents our input sequence. The rand scores quantify the extent to which a na\"ive K-means (N=3) clustering algorithm in different layers alone led to correct grouping of the input sentences.

\subsection{Sentence classification}
\subsubsection{Preparation of the data}
We used the PubMed corpus to fine-tune a sentence classification architecture. Class names and abbreviations are displayed in Table~\ref{tab:classes}. The corpus was supplied in pre-processed form, comprising 24,668 abstracts. For more information about the original dataset we refer to its original publication \cite{jin18}. Because of the PICO framework, methods for systematic review semi(automation) commonly focus on P, I, and O detection. A, M, R, and C classes are an additional feature of this corpus. They were included in the following experiment because they represent important information in abstracts and they occur in a vast majority of published trial text. Their exclusion can lead to false classification of sentences in full abstracts. In a preliminary experiment we summarized A, M, R, and C sentences as a generic class named ’Other’ in order to shift the model's focus to PIO classes. This resulted in high class imbalance, inferior classification scores and a loss of ability to predict these classes when supporting systematic reviewers during the screening process.  

In the following, abstracts that did not include a P, I, and O label were excluded. This left a total of 129,095 sentences for training, and 14,344 for testing (90:10 split). 
\subsubsection{Fine-tuning}
We carried out fine-tuning for sentence classification based on BERT-base (uncased), multilingual BERT (cased), and on SCIBERT. We changed the classification layer on top of the original BERT model. It remains as linear, fully connected layer but now employs the sigmoid cross-entropy loss with logits function for optimization. During training, this layer is optimised for predicting probabilities over all seven possible sentence labels. Therefore, this architecture enables multi-class, multi-label predictions. In comparison, the original BERT fine-tuning approach for sentence classification employed a softmax layer in order to obtain multi-class, single-label predictions of the most probable class only. During the training process the model then predicts class labels from Table 1 for each sentence. After each training step, backpropagation then adjusts the model's internal weights.       
%2 hours for multiling model finetuning. 
To save GPU resources, a maximal sequence length of 64, batch size 32, learning rate of $2\times10^{-5}$, a warm-up proportion of 0.1 and two epochs for training were used. 
\subsubsection{Post-training assignment of classes}
In the scope of the experiments for this paper, the model returns probabilities for the assignment of each class for every sentence. These probabilities were used to show effects of different probability thresholds (or simply assignment to the most probable class) on recall, precision and F1 scores. The number of classes was set to 7, thereby making use of the full PubMed dataset. 

%Then, the two datasets were used separately and in combined form. Additionally, instead of mixing sentences an alternative approach of combining the two datasets was tested. BERT layers can be excluded, or `frozen', by specifying trainable variables. First, only layers 7, 8 and embedding plus classification were fine tuned on the additional training data. Then, layers 7 and 8 were frozen and fine-tuning of all other layers was carried out for one more epoch. Evaluation was carried out with held-out data from both datasets, but showed no benefit over the other models. 
\subsection{Question answering}
\subsubsection{Preparation of the data}
Both the training and testing subsets from the ebm-nlp data were adapted to fit the SQuAD format. We merged both datasets in order to train a model which firstly correctly answers PICO questions on the basis of being trained with labelled ebm-nlp data, and secondly retains the flexibility of general-purpose question answering on the basis of SQuAD. We created sets of general, differently phrased P, I, and O questions for the purpose of training a broad representation of each PICO element question.

\begin{figure}[ht]
    \centering
    \begin{minipage}[ht]{0.48\textwidth}
\includegraphics[width=\linewidth]{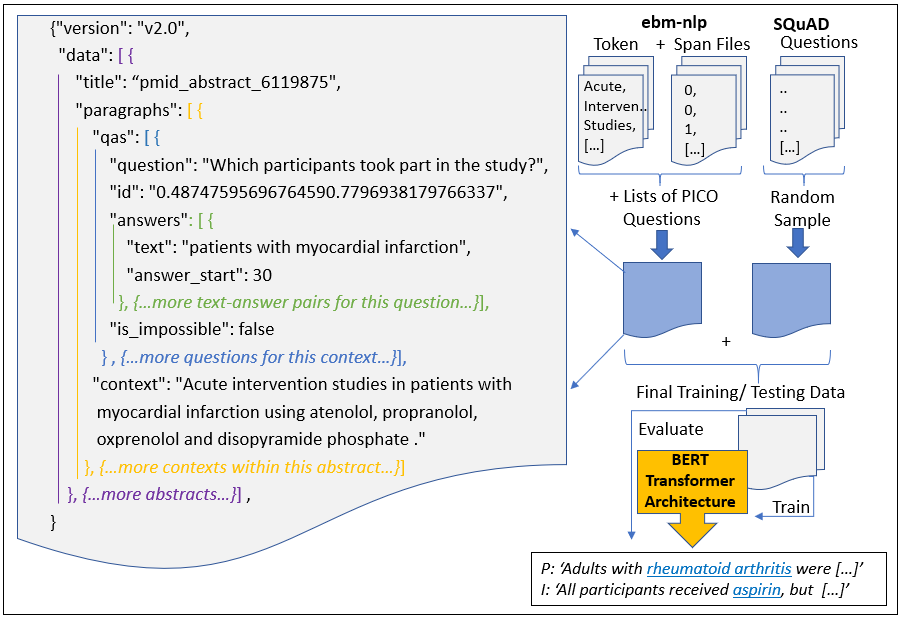}
\caption{Colour coded example for a population entity annotation, converted to SQuAD v.2 format. Combined data are used to train and evaluate the system.}\label{fig:squadExample}
\end{minipage}
\end{figure}

In this section we describe the process of adapting the ebm-nlp data to the second version of the SQuAD format, and then augmenting the training data with some of the original SQuAD data. Figure~\ref{fig:squadExample} shows an example of the converted data, together with a high-level software architecture description for our QA-BERT model. We created a conversion script to automate this task. To reduce context length, it first split each ebm-nlp abstract into sentences. For each P, I, and O class it checked the presence of annotated  entity spans in the ebm-nlp source files. Then, a question was randomly drawn from our set of general questions for this class, to complete a context and a span-answer pair in forming a new SQuAD-like question element. In cases where a sentence did not contain a span, a question was still chosen, but the answer was marked as impossible, with the plausible answer span set to begin at character 0. In the absence of impossible answers, the model would always return some part of the context as answer, and hence be of no use for rarer entities such as P, which only occurs in only 30\% of all context sentences. 

For the training data, each context can contain one possible answer, whereas for testing multiple question-answer pairs are permitted. An abstract is represented as a domain, subsuming its sentences and question answer-text pairs. In this format, our adapted data are compatible with the original SQuAD v.2 dataset, so we chose varying numbers of original SQuAD items and shuffled them into the training data. This augmentation of the training data aims to reduce the dependency on large labelled corpora for PICO entity extraction. Testing data can optionally be enriched in the same way, but for the presentation of our results we aimed to be comparable with previously published models and therefore chose to evaluate only on the subset of expert-annotated ebm-nlp testing data. 

\subsubsection{Fine-tuning}

The python Huggingface Transformers library was used for fine-tuning the question-answering models. This classification works by adding a span-classification head on top of a pre-trained transformer model. The span-classification mechanism learns to predict the most probable start and end positions of potential answers within a given context \cite{Wo19}.  

The Transformers library offers classes for tokenizers, BERT and other transformer models and provides methods for feature representation and optimization. We used BertForQuestionAnswering. Training  was  carried  out  on Google's Colab, using the GPU runtime option. We used a batch size of 18 per GPU and a learning rate of $3^{-5}$. Training lasted for 2 epochs, context length was limited to 150. To reduce the time needed to train, we only used BERT-base (uncased) weights as starting points, and used a maximum of 200 out of the 442 SQuAD domains.\\To date, the Transformers library includes several BERT, XLM, XLNet, DistilBERT and ALBERT  question answering models that can be fine-tuned with the scripts and data that we describe in this paper. 

\section{\uppercase{Results}}
\subsection{Feature representation and contextualization}
\begin{figure}[H]
\centering
\begin{minipage}[H]{0.465\textwidth}
\includegraphics[width=\linewidth]{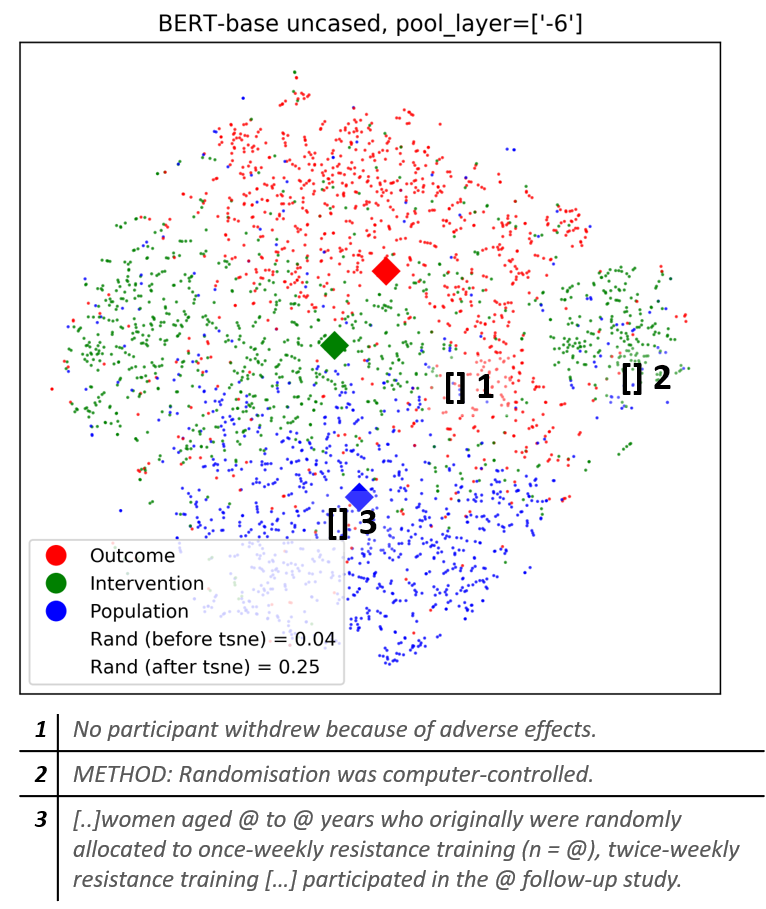}
\caption{Visualization of training sentences using BERT-base. The x and y-axis represent the two most dominant dimensions in the hidden state output, as selected by the t-SNE algorithm. This visualization uses the sixth layer from the top, and shows three examples of labelled P sentences and their embedded positions. }\label{fig:annotLayer}
\end{minipage}

\vspace{0.5 cm}
\begin{minipage}[b]{0.465\textwidth}
\includegraphics[width=\linewidth]{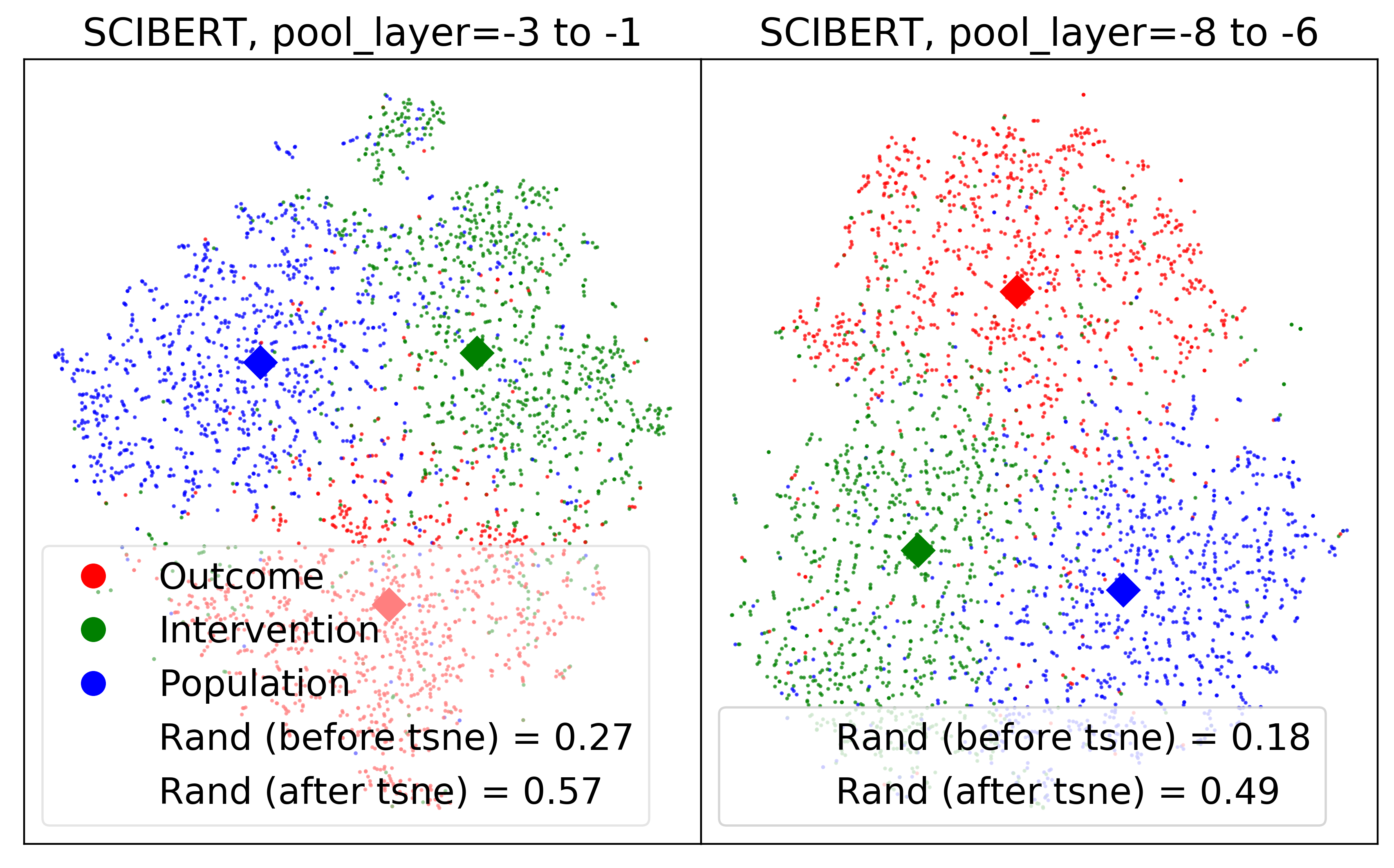}
\includegraphics[width=\linewidth]{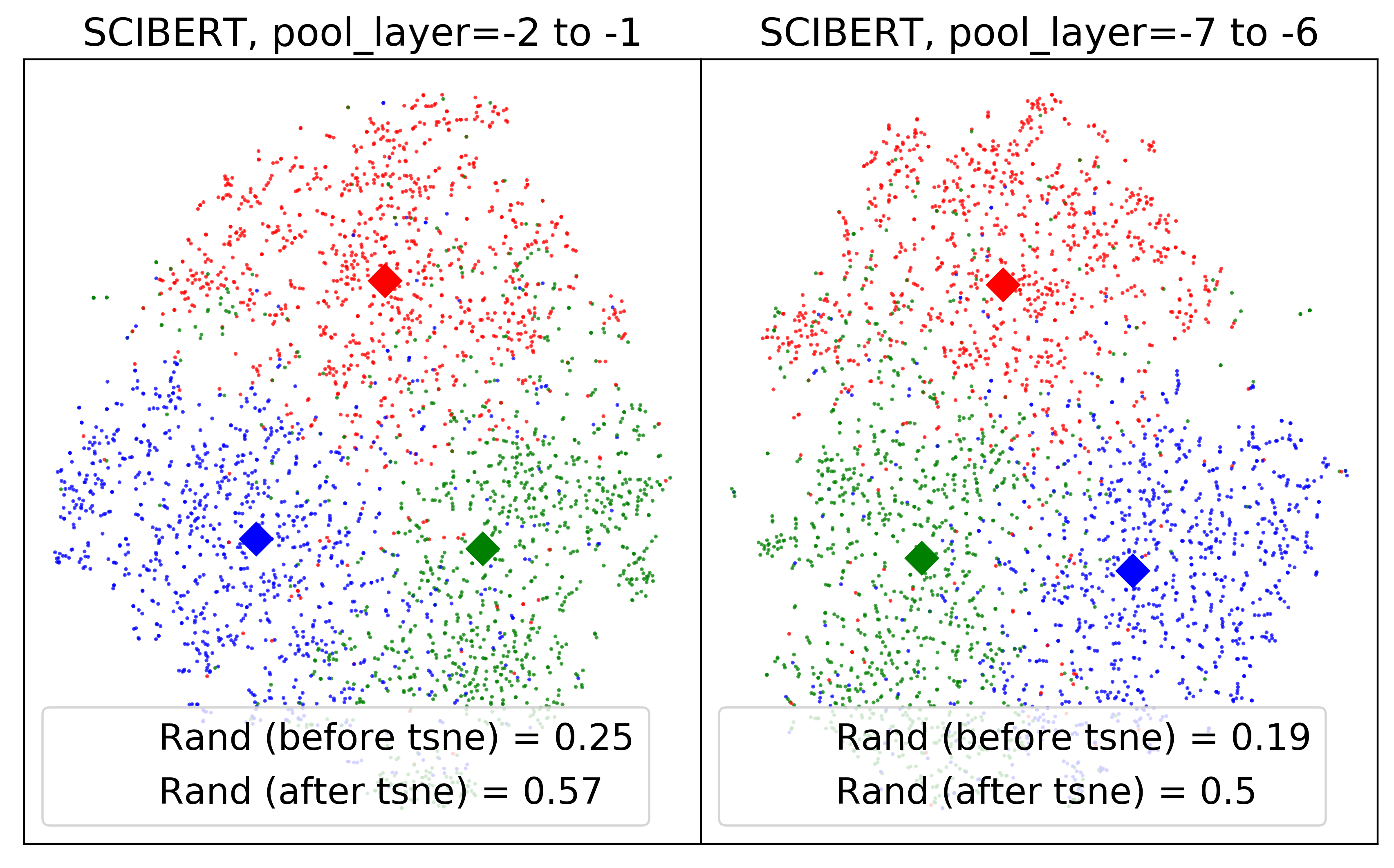}
\caption{Visualisation of training sentences using SCIBERT. The x and y-axes represent the two most dominant t-SNE reduced dimensions for each concatenation of layers}\label{fig:scibertEmb}
\end{minipage}
\end{figure}

Figure~\ref{fig:annotLayer} shows the dimensionality-reduced vectors for 3000 sentences in BERT-base, along with the positions of three exemplary sentences. All three examples were labelled as 'P' in the gold standard. 
This visualization highlights overlaps between the sentence data and ambiguity or noise in the labels. 
\begin{table*}[t]
\caption{Summary of results for the sentence classification. task}\label{tab:lstmResults}
\centering
\small
\begin{tabular}{cc}

\begin{tabular}{l|l|l||l|l||l|l}
\hline
\small
Tag & Model & Precision, Recall, F1&Model & Prec., Recall, F1& Model & Prec., Recall, F1\\\hline\hline
\multicolumn{7}{c}{Single-label case}\\\hline
P &LSTM &  0.89, 0.83, 0.86&BERT-base&0.92, 0.87, 0.89&SCIBERT&\textbf{0.92, 0.88, \underline{0.90}}\\\hline
I&&   0.75, 0.82, 0.78&&0.89, 0.88, 0.89&&\textbf{0.90, 0.88, \underline{0.89}}\\\hline
O&  &  0.84, 0.83, 0.84&&0.88, 0.93, 0.90&&\textbf{0.89, 0.94, \underline{0.92}}\\\hline
A& &\textbf{0.98, \underline{0.98}, \underline{098}}&&0.93, 0.94, 0.93&&0.94, 0.95, 0.95\\\hline
M&&0.87, 0.84, 0.86&&0.95, 0.93, 0.94&&\textbf{0.96, 0.93, \underline{0.95}}\\\hline
R&&\textbf{0.93, 0.96, \underline{0.95}}&&0.90, 0.94, 0.92&&0.91, 0.94, 0.92\\\hline
C&&\textbf{0.94, 0.91, \underline{0.92}}&&0.90, 0.83, 0.86&&0.89, 0.83, 0.86\\\hline\hline
\multicolumn{7}{c}{Multi-label case}\\\hline
P&Multilingual&0.87, 0.90, \textbf{0.88}&SCIBERT& 0.81, \textbf{\underline{0.93}}, 0.87  &SCIBERT&\textbf{\underline{0.97}}, 0.78, 0.87\\\hline
I&Thresh.: 0.3& 0.85, 0.90, \textbf{0.88}&0.2&  0.83, \textbf{\underline{0.92}}, 0.87  &0.8&\textbf{\underline{0.95}}, 0.75, 0.84\\\hline
O&&             0.87, 0.93, \textbf{0.90}       && 0.84, \textbf{\underline{0.95}}, 0.89  &&\textbf{\underline{0.95}}, 0.83, 0.89\\\hline
A&&             0.91, 0.94, 0.93                && 0.88, \textbf{0.96}, 0.92 &&\textbf{\underline{0.98}}, 0.91, \textbf{0.94}\\\hline
M&&             0.95, 0.94, 0.94                && 0.91, \textbf{\underline{0.95}}, 0.93  &&\textbf{\underline{0.99}}, 0.90, \textbf{0.94}\\\hline
R&&             0.88, 0.96, \textbf{0.92}       &&  0.86, \textbf{\underline{0.97}}, 0.91 &&\textbf{\underline{0.94}}, 0.85, 0.89 \\\hline
C&&             0.84, 0.87, \textbf{0.86}       && 0.79, \textbf{\underline{0.90}}, 0.84  &&\textbf{\underline{0.95}}, 0.70, 0.81\\\hline\hline
\multicolumn{7}{c}{Effect of threshold on metrics}\\\hline
\multicolumn{7}{c}{\includegraphics[width=0.9\linewidth]{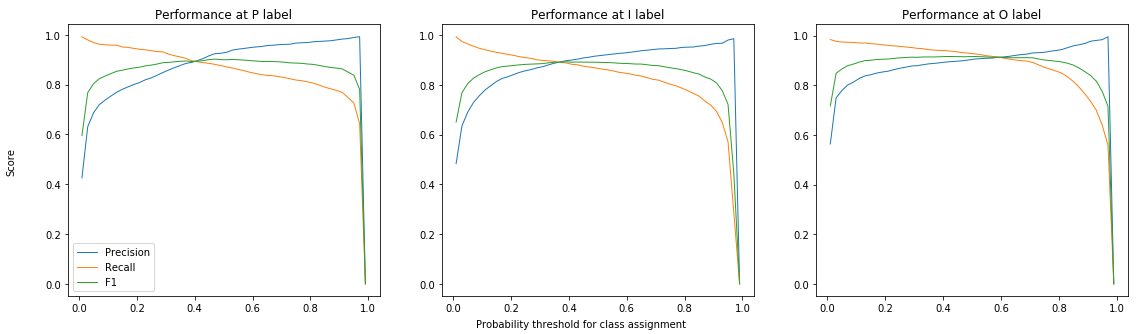}}\\\hline
\end{tabular} & 

\end{tabular}

\end{table*}

\begin{CJK*}{UTF8}{bsmi}

\begin{table*}[ht]
\caption{Predicting PICOs in Chinese and German. Classes were assigned based on foreign language inputs only. For reference, translations were provided by native speakers.}\label{tab:multilang}
\centering
\small
\begin{tabular}{cc}
\begin{tabular}{l|l}
\hline

 Prediction & Original sentences with English translations for reference \\\hline\hline
\multicolumn{2}{c}{Chinese}\\\hline

\rowcolor{apr!30}

Population & "方法:選擇2004-03/2005-03在惠州市第二人民醫院精神科精神分裂癥住院的患者60例,\\

&簡明精神癥狀量表總分＞30分,陰性癥狀量表總分＞35分."\cite{hua05}\\
& Translation: "Methods: In the Huizhou No. 2 People's hospital (Mar 2004 - Mar 2005), 60 patients   \\
&with psychiatric schizophrenia were selected. Total score of the Brief Psychiatric Rating Scale was  \\
& \textgreater 30, and total score of the Negative Syndrome Scale was \textgreater 35 in each patient."\\\hline
\rowcolor{cya!70}
 Intervention &"1 隨機分為2組,泰必利組與奎的平組,每組30例,患者家屬知情同意.\\
 & Translation: "1. They were randomly divided into 2 groups, Tiapride group and Quetiapine group, with \\ 
 & 30 cases in each group. Patients' family was informed and their consent was obtained."\\\hline

&初始劑量25 mg,早晚各1次,以后隔日增加50 mg,每日劑量范圍300～550 mg."\cite{hua05}\\
&Translation: "The initial dose was 25 mg, once in the morning and evening. The dose was increased by \\
&50 mg every other day, reaching a daily dose in the range of 300 and 550 mg." \\
\rowcolor{gre!40}
Outcome &"3 效果評估:使用陰性癥狀量表評定用藥前后療效及陰性癥狀改善情況,使用副反應量表評價藥\\
& 物的安全性."\cite{hua05}\\
& Translation: "3 Evaluation: the Negative Symptom Scale was used to evaluate the efficacy of the drug \\
& before and after treatment and the improvement of the negative symptoms. The Treatment Emergent \\
&Symptom Scale was used to evaluate the safety of the drug."\\\hline
\rowcolor{lav!50}
Method & "實驗過程為雙盲."\cite{hua05}\\
& Translation: "The experimental process was double-blind."\\\hline

\rowcolor{whi}
\multicolumn{2}{c}{German}\\\hline
\rowcolor{yel!50}
Aim & "Ziel: Untersuchung der Wirksamkeit ambulanten Heilfastens auf Schmerz, Befindlichkeit und  \\
&Gelenkfunktion bei Patienten mit Arthrose."\cite{sch10}\\
&Translation: "Aim: to investigate outpatient therapeutic fasting and its effects on pain, wellbeing and \\
&joint-function in patients with osteoarthritis."\\\hline
\rowcolor{lav!50}
Method & "Patienten und Methoden: Prospektive, unkontrollierte Pilotstudie."\cite{sch10}\\
& Translation: "Patients and methods: prospective, uncontrolled pilot study"\\\hline
\rowcolor{gre!40}
Outcomes& "Anlauf-, Belastungs-, Ruheschmerz (VAS); Druckschmerzschwelle (DSS); [..] \cite{sch10}\\
& Translation: "Pain was measured during warm-up, stress and resting (VAS); Onset of pain under \\
& pressure (DSS);[..]"\\\hline
\rowcolor{red!40}
Conclusion & "Schlussfolgerung: Heilfasten unter Ärztlicher Aufsicht kann die Symptomatik bei Patienten mit  \\

& moderater Arthrose positiv beeinflussen." \cite{sch10}\\
& Translation: "Conclusions: therapeutic fasting, under supervision of doctors, can have a positive \\
&effect on the symptoms of patients with moderate osteoarthritis."
\end{tabular} & 

\end{tabular}

\end{table*}

\end{CJK*}
Sentences 1 and 2 are labelled incorrectly, and clearly appear far away from the population class centroid. Sentence 3 is an example of an ambiguous case. It appears very close to the population centroid, but neither its label nor its position reflect the intervention content. This supports a need for multiple tags per sentence, and the fine-tuning of weights within the network.
%\vfill

 Figure~\ref{fig:scibertEmb} shows the same set of sentences, represented by concatenations of SCIBERT outputs. SCIBERT was chosen as an additional baseline model for fine-tuning because it provided the best representation of embedded PICO sentences. When clustered, its embeddings yielded an adjusted rand score of 0.57 for a concatenation of the two layers, compared with 0.25 for BERT-base.

\subsection{Sentence classification}
Precision, recall, and F1 scores, including a comparison with the LSTM, are summarized in Table \ref{tab:lstmResults}. Underlined scores represent the top score across all models, and scores in bold are the best results for single- and multi-label cases respectively. The LSTM assigns one label only and was outperformed in all classes of main interest (P, I, and O).      

A potential pitfall of turning this task into multi-label classification is an increase of false-positive predictions, as more labels are assigned than given in the single-labelled testing data in the first place. However, the fine-tuned BERT models achieved high F1 scores, and large improvements in terms of recall and precision. In its last row, Table~\ref{tab:lstmResults} shows different probability thresholds for class assignment when using the PubMed dataset and our fine-tuned SCIBERT model for multi-label prediction. After obtaining the model's predictions, a simple threshold parameter can be used to obtain the final class labels. On our labelled testing data, we tested 50 evenly spaced thresholds between 0 and 1 in order to obtain these graphs. Here, recall and precision scores in ranges between 0.92 and 0.97 are possible with F1 scores not dropping below 0.84 for the main classes of interest. In practice, the detachment between model predictions and assignment of labels means that a reviewer who wishes to switch between high recall and high precision results can do so very quickly, without obtaining new predictions from the model itself.  

More visualizations can be found in this project's GitHub repository \footnote{https://github.com/L-ENA/HealthINF2020}, including true class labels and a detailed breakdown of true and false predictions for each class. The highest proportion of false classification appears between the results and conclusion classes.

The fine-tuned multilingual model showed marginally inferior classification scores on the exclusively English testing data. However, this model's contribution is not limited to the English language because its interior weights embed a shared vocabulary of 100 languages, including German and Chinese\footnote{For a full list see \url{https://github.com/google-research/bert/blob/master/multilingual.md}}. Our evaluation of the multilingual model's capacity for language transfer is of a qualitative nature, as there were no labelled Chinese or German data available. Table~\ref{tab:multilang} shows examples of two abstracts, as predicted by the model. Additionally, this table demonstrates how a sentence prediction model can be used to highlight text. With the current infrastructure it is possible to highlight PICOs selectively, to highlight all classes simultaneously, and to adjust thresholds for class assignment in order to increase or decrease the amount of highlighted sentences. When applied to full texts of RCTs and cohort studies, we found that the model retained its ability to identify and highlight key sentences correctly for each class. \\~
 
We tested various report types, as well as recent and old publications, but remain cautious that large scale testing on labelled data is needed to draw solid conclusions on these model's abilities for transfer learning. For further examples in the English language, we refer to our GitHub repository.

\vfill

\subsection{Question answering}
We trained and evaluated a model for each P, I, and O class. Table~\ref{tab:qa} shows our results, indicated as QA-BERT, compared with the currently published leader board for the ebm-nlp data \cite{ebm19} and results reported by the authors of SCIBERT \cite{Bel19}. For the P and I classes, our models outperformed the results on this leader board. The index in our model names indicates the amount of additional SQuAD domains added to the training data. We never used the full SQuAD data in order to reduce time for training but observed increased performance when adding additional data. For classifying I entities, an increase from 20 to 200 additional SQuAD domains resulted in an increase of 8\% for the F1 score, whereas the increase for the O domain was less than 1\%. After training a model with 200 additional SQuAD domains, we also evaluated it on the original SQuAD development set and obtained a F1 score of 0.72 for this general reading comprehension task.  

In this evaluation, the F1 scores represent the overlap of labelled and predicted answer spans on token level. We also obtained scores for the subgroups of sentences that did not contain an answer versus the ones that actually included PICO elements. These results are shown in Table~\ref{tab:subgroups}. 

For the P class, only 30\% of all sentences included an entity, whereas its sub-classes age, gender, condition and size averaged 10\% each. In the remaining classes, these percentages were higher. F1 scores for correctly detecting that a sentence includes no PICO element exceeded 0.92 in all classes. This indicates that the addition of impossible answer elements was successful, and that the model learned a representation of how to discriminate PICO contexts. The scores for correctly predicting PICOs in positive scenarios are lower. These results are presented in Table~\ref{tab:subgroups}. Here, two factors could influence this score in a negative way. First, labelled spans can be noisy. Training spans were annotated by crowd workers and the authors of the original dataset noted inter-annotator disagreement. Often, these spans include full stops, other punctuation or different levels of detail describing a PICO. The F1 score decreases if the model predicts a PICO, but the predicted span includes marginal differences that were not marked up by the experts who annotated the testing set. Second, some spans include multiple PICOs, sometimes across sentence boundaries. Other spans mark up single PICOS in succession. In these cases the model might find multiple PICOs in a row, and annotate them as one or vice versa.    

\begin{table}
\caption{Question Answering versus entity recognition \\results.}
    \label{tab:qa}
    \centering
    \begin{tabular}{l|l|l}
    \hline
    Class & Model & F1 score\\\hline\hline
     P      &lstm-crf$_0$	    & 0.78 \\\hline
            &lstm-crf-BERT$_0$  & 0.68\\\hline
            & QA-BERT$_{20}$           & \underline{0.87}\\\hline\hline
    I       &  lstm-crf-pattern$_1$ & 0.7     \\  \hline
           &    lstm-crf-BERT$_0$& 0.57 \\\hline
           &   QA-BERT$_{20}$          &  0.67\\\hline
           &   QA-BERT$_{200}$          &  \underline{0.75}\\\hline\hline
    O       &  lstm-crf-BERT$_0$ & \underline{0.78} \\\hline
          
           & QA-BERT$_{100}$ &  0.67\\  \hline\hline  
    PIO   & SCIBERT mean  & 0.72 \\ \hline
    & QA-BERT mean& \underline{0.75}\\\hline \hline
    \multicolumn{3}{c}{P subclasses when annotated:}\\
    \multicolumn{3}{c}{Age, Gender, Condition, Size}\\\hline\hline
     All & lstm-crf$_0$ & $0.4$\\\hline
     All & QA-BERT$_{20}$ & \underline{0.53}\\\hline
    \multicolumn{3}{c}{Performance in general SQuAD task}\\\hline\hline 
    All & QA-BERT$_{200}$ & 0.72\\\hline
    \end{tabular}
    
\end{table}

\begin{table*}[t]
\caption{Subgroups of possible sentences versus impossible sentences.}
    \label{tab:subgroups}
    \centering
    
    \begin{tabular}{l|l|l|l|l}
    \hline
    Class&Model&\%&F1&F1\\
    &&Possible&When Possible&When Impossible\\\hline\hline
    
     P  & QA-BERT$_{20}$ & 30\% & 0.74 & 0.92\\\hline
     I  & QA-BERT$_{200}$ & 53\% & 0.60& 0.94\\\hline
     0 &  QA-BERT$_{100}$ &60\%   & 0.52 & 0.92\\\hline
     P$_{all}$ &QA-BERT$_{20}$&10\% & 0.53 & 0.97\\\hline
    \end{tabular}
    
\end{table*}

\begin{table}
\caption{This table shows two examples for intervention span predictions in QA-BERT$_{200}$. On the official SQuAD development set, the same model achieved a good score, an exemplary question and prediction for this is given in the bottom row.}\label{fig:squadSpans} \centering
\begin{tabular}{p{1.45cm}|p{5cm}} \hline
\multicolumn{2}{c}{Which intervention did the participants receive?}\\\hline\hline
  Prediction & Context \\\hline
  
      auditory integrative training & To explore the short-term treatment effect of the auditory integrative training on autistic children and provide them with clinical support for rehabilitative treatment. \\\hline
     ketorolac trometha- mine 10 mg and 30 mg suppositories & The analgesia activity of ketorolac tromethamine 10 mg and 30 mg suppositories were evaluated after single dose administration by assessing pain intensity and pain relief using a 4 point scale ( VRS ).\\\hline\hline
     \multicolumn{2}{c}{What do power station steam turbines use as a}\\
     \multicolumn{2}{c}{cold sink in the absence of CHP?}\\\hline
      
     surface condensers & Where CHP is not used, steam turbines in power stations use surface condensers as a cold sink. The condensers are cooled by water flow from oceans, rivers, lakes, and often by cooling towers […]\\\hline

\end{tabular}
\end{table}

\section{\uppercase{Discussion}}

In this work, we have shown possibilities for sentence classification and data extraction of PICO characteristics from abstracts of RCTs.

For sentence classification, models based on transformers can predict multiple labels per sentence, even if trained on a corpus that assigns a single label only. Additionally, these architectures show a great level of flexibility with respect to adjusting precision and recall scores. Recall is an important metric in SR tasks and the architectures proposed in this paper enable a post-classification trade-off setting that can be adjusted in the process of supporting reviewers in real-world reviewing tasks. 

However, tagging whole sentences with respect to populations, interventions and outcomes might not be an ideal method to advance systematic review automation. Identifying a sentence's tag could be helpful for highlighting abstracts from literature searches. This focuses the reader's attention on sentences, but is less helpful for automatically determining whether a specific entity (e.g. the drug aspirin) is mentioned.

Our implementation of the question answering task has shown that a substantial amount of PICO entities can be identified in abstracts on a token level. This is an important step towards reliable systematic review automation. With our provided code and data, the QA-BERT model can be switched with more advanced transformer architectures, including XLM, XLNet, DistilBERT and ALBERT pre-trained models. More detailed investigations into multilingual predictions \cite{clef20} pre-processing and predicting more than one PICO per sentence are reserved for future work.

\subsection{Limitations}

Limitations in the automatically annotated PubMed training data mostly consist of incomplete detection or noise P, I, and O entities due to the single labelling. We did not have access to multilingual annotated PICO corpora for testing, and therefore tested the model on German abstracts found on PubMed, as well as Chinese data provided by the Cochrane Schizophrenia Group.   

For the question answering, we limited the use of original SQuAD domains to enrich our data. This was done in order to save computing resources, as an addition of 100 SQuAD domains resulted in training time increases of two hours, depending on various other parameter settings. Adjusted parameters include increased batch size, and decreased maximal context length in order to reduce training time.

%The original SQuAD v.2 dataset contains spans for plausible answers whenever an answer is marked as impossible. These are subsequently used in the evaluation process, in order to measure where the model wrongly predicted these answers. The ebm-nlp corpus did not contain alternative span data. We simply marked the first two words of each sentence as possible answer, but acknowledge that manually crafted rules or simple heuristics could lead to a better annotation.
\section{\uppercase{Conclusion}}
With this paper we aimed to explore state-of-the-art NLP methods to advance systematic review (semi)automation. Both of the presented fine-tuning approaches for transformers demonstrated flexibility and high performance. We contributed an approach to deal with ambiguity in whole-sentence predictions, and proposed the usage of a completely different approach to entity recognition in settings where training data are sparse.    

In conclusion we wish to emphasize our argument that for future applications, interoperability is important. Instead of developing yet another stand-alone organizational interface with a machine learning classifier that works on limited data only, the focus should be to develop and train cross-domain and neural models that can be integrated into the backend of existing platforms. The performance of these models should be comparable on standardized datasets, evaluation scripts and leader boards.

The logical next step, which remains less explored in the current literature because of its complexity, is the task of predicting an RCT's included or excluded status on the basis of PICOs identified in its text. For this task, more complex architectures that include drug or intervention ontologies could be integrated. Additionally, information from already completed reviews could be re-used as training data.

\section*{\uppercase{Acknowledgements}}
We would like to thank Clive Adams for providing testing data and feedback for this project. We thank Vincent Cheng for the Chinese translation. Furthermore, we thank the BERT team at Google Research and Allenai for making their pre-trained model weights available. Finally, we acknowledge the Huggingface team and thank them for implementing the SQuAD classes for Transformers.

\section*{\uppercase{Funding}}
LS was funded by the National Institute for Health Research (NIHR Systematic Review Fellowship, RM-SR-2017-09-028). The views expressed in this publication are those of the author(s) and not necessarily those of the NHS, the NIHR or the Department of Health and Social Care.

\bibliographystyle{apalike}
{\small
\bibliography{example}}

\section*{\uppercase{Appendix}}
\section*{Availability of the code and data}

Scripts and supplementary material, as well as further illustrations are available from \url{https://github.com/L-ENA/HealthINF2020}. Training data for sentence classification and question answering are freely available from the cited sources.

Additionally, the Cochrane Schizophrenia Group extracted, annotated and made available data from studies included in over 200 systematic reviews. This aims at supporting the development of methods for reviewing tasks, and to increase the re-use of their data. These data include risk-of-bias assessment, results including all clean and published outcome data extracted by reviewers, data on PICOs, methods, and identifiers such as PubMed ID and a link to their study-based register. Additionally, a senior reviewer recently carried out a manual analysis of all 33,000 outcome names in these reviews, parsed and allocated to 15,000 unique outcomes in eight main categories \cite{sch19}.

%\section*{Examples for SCIBERT predictions}
%\subsection{*Sentence classification}
%\begin{figure}[h]
%\begin{minipage}[t]{0.47\textwidth}
%\includegraphics[width=\linewidth]{figs/BertConfusion.png}
%\caption{Class predictions versus the true labels in the fine-tuned SCIBERT model}\label{fig:bertConfusion}
%\end{minipage}
%\end{figure}

%\onecolumn

%\begin{figure*}[h]
%\centering
%\begin{minipage}[t]{0.7\textwidth}
%\includegraphics[width=\linewidth]{figs/ScibertTable.png}
%\caption{Prediction examples from fine-tuned SCIBERT model. Rows 1,2, and 5 show cases where multi-label prediction correctly identified multiple classes. Rows 3, 4, and 6 show cases where the model correctly predicted ambiguous or wrongly labelled data. Rows 7 and 8 demonstrate wrong predictions}\label{t:preds}
%\end{minipage}
%\end{figure*}

%\begin{figure*}[h]
%\centering
%\begin{minipage}[t]{0.99\textwidth}
%\includegraphics[width=\linewidth]{figs/english.PNG}
%\caption{Highlighted sentences of an abstract, taken from a literature search}\label{t:preds}
%\end{minipage}
%\end{figure*}

%\twocolumn

%\ifCLASSOPTIONcaptionsoff
  %\newpage
%\fi

\end{document}